\documentclass[11pt]{article}
\usepackage{etoolbox}
\newtoggle{anonymous}
\newtoggle{arxiv}
\togglefalse{arxiv}
\iftoggle{arxiv}{%
    \usepackage{arxiv}
    \togglefalse{anonymous}
    \newcommand{\cvspace}[1]{}
}{%
    \togglefalse{anonymous}
    \usepackage[\iftoggle{anonymous}{review}{final}]{acl}
    \newcommand{\cvspace}[1]{}
}

\usepackage{times}
\usepackage{latexsym}
\usepackage[T1]{fontenc}
\usepackage[utf8]{inputenc}
\usepackage{microtype}
\usepackage{inconsolata}
\usepackage{graphicx}
\usepackage{hyperref}
\usepackage{url}
\usepackage{booktabs}
\usepackage{amsfonts}
\usepackage{nicefrac}
\usepackage{mathtools,bm,amssymb}
\usepackage{xcolor}
\usepackage{hyperref}
\usepackage{subcaption,placeins}
\usepackage{adjustbox,multirow,dcolumn}
\usepackage{algorithm,algpseudocode,setspace}
\usepackage{xstring}
\usepackage{cleveref}
\usepackage{tikz}
\usepackage{pgfplots}
\usepackage{comment}

\newcommand{\email}[1]{\href{mailto:#1}{\normalfont {\texttt{#1}}}}

\usepackage{amsmath,amsfonts,bm}

\def\1{\bm{1}}

\DeclareMathAlphabet{\mathsfit}{\encodingdefault}{\sfdefault}{m}{sl}
\SetMathAlphabet{\mathsfit}{bold}{\encodingdefault}{\sfdefault}{bx}{n}

\newcommand{\lr}{\alpha}

\DeclareMathOperator*{\argmin}{arg\,min}

\makeatletter
\DeclareRobustCommand\onedot{\futurelet\@let@token\@onedot}
\def\@onedot{\ifx\@let@token.\else.\null\fi}

\newcommand{\eg}{\emph{e.g\@\onedot}}
\newcommand{\etc}{\emph{etc\@\onedot}}
\newcommand{\etal}{\emph{et~al\@\onedot}}

\newcommand{\ie}{\emph{i.e\@\onedot}}
\newcommand{\versus}{\texorpdfstring{\emph{vs\@\onedot}}{vs.}}

\newcommand{\first}{1\textsuperscript{st}}
\newcommand{\second}{2\textsuperscript{nd}}

\newcommand{\ordinal}[1]{{#1}\textsuperscript{th}}

\graphicspath{{figures/}}

\usetikzlibrary{external}
\usepgfplotslibrary{colorbrewer}
\pgfplotsset{compat=newest}
\newcolumntype{d}{D{.}{.}{3.2}}
\makeatletter
\newcolumntype{B}{>{\boldmath\DC@{.}{.}{3.2}}c<{\DC@end}}
\makeatother

\newcommand{\tlbox}[1]{\begin{tabular}[c]{@{}l@{}}#1\end{tabular}}

\newcommand*\best[1]{{\textbf{#1}}}
\newcommand*\std[2]{{{#1}{\scriptsize $\pm${#2}}}}
\newcommand*\beststd[2]{{\textbf{#1}{\scriptsize $\pm${#2}}}}

\newcommand{\numfirstorder}{6}

\newcommand{\nummetrics}{8}
\newcommand{\numtrials}{3}

\DeclarePairedDelimiter{\parens}{\lparen}{\rparen}

\DeclarePairedDelimiter{\bracks}{[}{]}
\DeclarePairedDelimiter{\braces}{\{}{\}}
\DeclarePairedDelimiter{\verts}{\lvert}{\rvert}

\newcommand{\diff}[2]{\frac{\partial{#1}}{\partial{#2}}}

\newcommand{\x}{{\mathbf{x}}}

\newcommand{\density}{{\rho}}
\newcommand{\sval}{{\mathbf{s}}}
\newcommand{\weight}{{\bm\theta}}
\newcommand{\sparseweight}{\weight_\mathrm{sp}}
\newcommand{\mask}{{\bm\tau}}
\newcommand{\dataset}{\mathcal{D}}

\newcommand{\trainset}{\dataset_\textrm{train}}

\newcommand{\realset}{\mathbb{R}}

\DeclareMathOperator{\smetric}{\mathcal{S}}
\DeclareMathOperator{\topk}{\textrm{top}}

\DeclareMathOperator{\loss}{\mathcal{L}}

\DeclareMathOperator{\indicator}{\mathbf{1}}

\DeclareMathOperator{\minibatch}{minibatch}

\title{%
    \mbox{Refining Salience-Aware Sparse Fine-Tuning}
    \mbox{Strategies for Language Models}}
\author{%
    Xinxin Liu\textsuperscript{\bf 1,2} \\ \And
    Aaron Thomas\textsuperscript{\bf 3} \\ \And
    Cheng Zhang\textsuperscript{\bf 4} \\ \AND
    Jianyi Cheng\textsuperscript{\bf 5} \\ \And
    Yiren Zhao\textsuperscript{\bf 4} \\ \And
    Xitong Gao\textsuperscript{\bf 2,6}%
    \thanks{Corresponding author, \email{xt.gao@siat.ac.cn}.}
    \AND \vspace{-1em} \\
    \textsuperscript{\bf 1} Southern University of Science and Technology \\
    \textsuperscript{\bf 2} Shenzhen Institutes of Advanced Technology, CAS \\
    \textsuperscript{\bf 3} University of Birmingham \hspace{0.5em}
    \textsuperscript{\bf 4} Imperial College London \hspace{0.5em}
    \textsuperscript{\bf 5} University of Edinburgh \\
    \textsuperscript{\bf 6} Shenzhen University of Advanced Technology
}
\newcommand{\repourl}{%
    \iftoggle{anonymous}{%
        Anonymized for review.
    }{%
        Available at: \url{https://github.com/0-ml/speft}.
    }
}

\begin{document}

\maketitle

\begin{abstract}
    Parameter-Efficient Fine-Tuning (PEFT)
    has gained prominence
    through low-rank adaptation methods like LoRA.
    In this paper,
    we focus on sparsity-based PEFT (SPEFT),
    which introduces trainable sparse adaptations
    to the weight matrices in the model,
    offering greater flexibility
    in selecting fine-tuned parameters
    compared to low-rank methods.
    We conduct the first systematic evaluation
    of salience metrics for SPEFT,
    inspired by zero-cost NAS proxies,
    and identify simple gradient-based metrics
    is reliable,
    and results are on par with the best alternatives,
    offering both computational efficiency and robust performance.
    Additionally,
    we compare static and dynamic masking strategies,
    finding that static masking,
    which predetermines non-zero entries before training,
    delivers efficiency without sacrificing performance,
    while dynamic masking offers no substantial benefits.
    Across NLP tasks,
    a simple gradient-based, static SPEFT
    consistently outperforms other fine-tuning methods for LLMs,
    providing a simple yet effective baseline for SPEFT.
    Our work challenges the notion
    that complexity is necessary for effective PEFT,
    while our open-source framework
    establishes a reproducible benchmark
    for future research%
    \footnote{\repourl{}}.
\end{abstract}

\section{Introduction}\label{sec:intro}

Pretrained large language models (LLMs)
have demonstrated strong performance
across various natural language processing (NLP) tasks
\cite{brown2020language}.
A typical approach for adapting these LLMs
to specific downstream tasks
involves fine-tuning their trainable parameters.
However,
this process can be prohibitively expensive
on consumer-grade hardwares,
if we consider training all free parameters,
especially on LLMs exceeding a billion parameters.
For example,
models with over 100 billion parameters,
such as BLOOM,
required training with 384 GPUs
across 48 distributed computing nodes
\cite{luccioni2023estimating}.
Instead of training all parameters,
an alternative fine-tuning paradigm
that enables model training on new tasks
with minimal computational resources
is \emph{Parameter-Efficient Fine-Tuning} (PEFT).
This method aims to learn only a small set of parameters
in order to adapt the model to the new task,
substantially lowers the computational resource requirements
\cite{ansell2021composable, hu2021lora}.
\begin{figure*}[t]
	\centering%
    {\includegraphics[
        width=1\linewidth, trim=15pt 15pt 15pt 15pt,
    ]{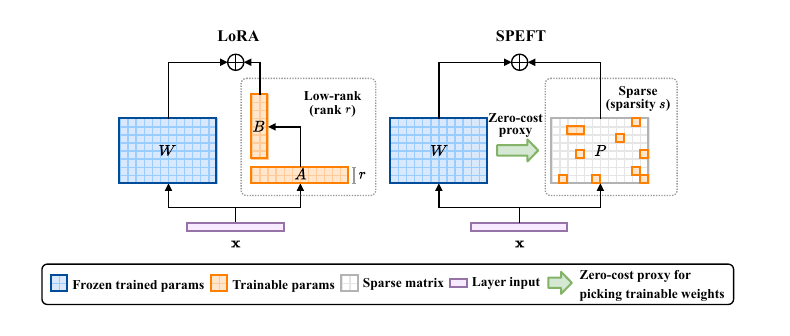}}
    \caption{%
		Comparison between LoRA~\cite{hu2021lora} and SPEFT\@.
		LoRA freezes pretrained weights \( \weight_0 \)
		and updates the low-rank terms \( A \) and \( B \),
		while SPEFT adopts zero-cost proxies
		to build a sparse adapter \( \sparseweight \),
		to update the weight elements
		that contribute most to the downstream task.
    }\label{fig:motivation}
\end{figure*}

Existing effort on PEFT methods mainly focuses on two categorizes,
low-rank-based and sparsity-based adaptation approaches.
LoRA \cite{hu2021lora},
a popular low-rank adaptation method
reparameterizes the model weight of each layer
(\( \weight \in \realset^{d_1 \times d_2} \))
as \( \weight \triangleq \weight_0 + BA \),
where \( \weight_0 \) denotes the pretrained weight matrix
which remains fixed during fine-tuning,
\( B \in \realset^{d_1 \times r} \)
and \( A \in \realset^{r \times d_2} \)
are trainable weights of a lower rank
with \( r \ll \min\{d_1, d_2\} \).
Recently,
sparsity-based PEFT (SPEFT)
has emerged as an alternative approach
which constructs an alternate reparameterization,
\( \weight \triangleq \weight_0 + \sparseweight \),
where \( \sparseweight \) is an extremely sparse matrix,
and updates solely its non-zero entries.
\Cref{fig:motivation}
illustrates the distinction
between the two categories of PEFT methods.
Previous sparse PEFT methods
\cite{
	guo2020parameter,
    sung2021fishmask,
    ansell2021composable}
have employed various first- and second-order metrics
for determining these non-zero entries
and adopted distinct approaches
for handling the sparsity mask during training.
The varying constructions and training-time treatments
of the sparsity mask
lead us to the following research questions
on the basic design principles for SPEFT\@:
\begin{itemize}\it
	\item Which salience metric or proxy
    is optimal for determining a sparsity mask?

	\item Is a static mask determined
    prior to the start of training sufficient,
	or is a dynamically updated pruning mask preferable?
\end{itemize}

In this paper,
we systematically re-examine the design principles for SPEFT
and conduct an evaluation
across distinct salience metrics.
Drawing inspiration from recent advancements
in zero-cost Network Architecture Search (NAS) proxies,
which explore diverse low-cost proxies
for determining parameter importance
that has incorporated both first-order
(\eg{}, weight magnitude, gradients, SNIP \cite{snip}, \etc{})
and second-order estimators
(\eg{}, GRaSP \cite{wang2020grasp},
Fisher information \cite{sung2021fishmask}, \etc{}),
we discovered that these NAS proxies
encompasses many salience metrics used in SPEFT
for sparsity mask construction
(DiffPruning \cite{guo2020parameter},
FishMASK \cite{sung2021fishmask},
\etc{}).
Consequently,
inspired by recent zero-cost NAS metrics
that have shown strong performance
to construct sparsity masks,
we are the first
to comprehensively evaluate \nummetrics{} different salience metrics
in the context of SPEFT for LLMs.
Furthermore,
we investigate both dynamic and static masking approaches,
where a dynamic mask matrix \( \mask \)
changes during training,
while a static mask
maintains a static \( \mask \) binary matrix
throughout the PEFT process.
We make the following contributions:
\begin{itemize}
	\item
    We systematically evaluate \nummetrics{}
    different salience metrics
    for constructing sparsity masks in SPEFT
    and empirically show that gradient-based SPEFT
    offers strong performance,
    while second-order metrics,
    such as Fisher information,
    do not significantly enhance SPEFT performance.

    \item
    We found that dynamic masking strategies
    do not surpass the effectiveness
    of a simple static mask
    predefined before training in SPEFT\@.
    This approach affords greater acceleration opportunities,
    as fixed indices are predetermined
    and this avoids the mask re-computation cost.

    \item
    Our results indicate that a simple gradient-based,
    static SPEFT method
    delivers the best trade-off
    between effectiveness and efficiency.
    For instance,
    for RoBERTa-base \cite{liu2019roberta}
    on MRPC \cite{dolan-brockett2005mrpc} task,
    our method achieves 0.98\% higher
    than the baseline
    given the same amount of trainable parameters.
    Gradient-based SPEFT outperforms LoRA
    by 22.6\% on GSM8k \cite{cobbe2021gsm8k}
    when trained on MetaMathQA \cite{yu2024metamath}.
    Consequently,
    we advocate for this SPEFT variant
    to be considered a strong baseline
    for subsequent developments in this field.
\end{itemize}


\section{Related Work}\label{sec:related}

\subsection{PEFT Methods}

With the advent of large language models,
fine-tuning these models on downstream tasks
can be prohibitively expensive
due to the sheer number of trainable parameters.
A suite of parameter-efficient fine-tuning (PEFT) methods
have been proposed to address this issue.

\textbf{Low-rank adaptation} \cite{hu2021lora}
is a popular method in PEFT
which reparameterizes the weight matrix
of each layer
(\( \weight \in \realset^{d_1 \times d_2} \))
as \( \weight = \weight_0 + BA \).
Here,
\( \weight_0 \in \realset^{d_1 \times d_2} \)
is the pretrained weight matrix,
and \( B \in \realset^{d_1 \times r} \)
and \( A \in \realset^{r \times d_2} \)
are lower-rank matrices
with \( r \ll \min(d_1, d_2) \).
By making only \( A \) and \( B \) trainable,
this method significantly reduces the number
of trainable parameters,
thereby lowering computational resource requirements.
LoRA has demonstrated effectiveness
in reducing trainable parameters
for fine-tuning large language models,
while maintaining strong fine-tuned performance
across various downstream tasks
compared to full fine-tuning.

\textbf{Sparsity-based adaptation}
Since the advent of low-rank adaptation,
sparsity-based adaptation has emerged
as an alternative approach to PEFT\@.
It constructs sparse trainable matrix
\( \sparseweight \) reparameterization
for each layer weight
\( \weight = \weight_0 + \sparseweight \),
where \(
    \verts{\sparseweight}_0
    \leq s \ll d_1 \times d_2
\),
and \( s \) represents the number of non-zero entries.
The gradient updates
only happen to the non-zero entries of the sparse matrices
during fine-tuning.
Since the sparse matrix \( \sparseweight \)
is typically constructed to be extremely sparse,
this approach can also achieve notable parameter efficiency,
and the sparsity masking strategy plays a crucial role
in determining impactful trainable parameters
for fine-tuning.

This approach has been explored
in various forms in the literature.
Earlier works
such as DiffPruning \cite{guo2020parameter}
learns a sparsity mask
with straight-through gradient estimator
\cite{bengio2013estimating,hubara2016binarized}
to select important parameters
for downstream tasks.
FishMASK \cite{sung2021fishmask}
applies a static sparsity mask
from training outset,
guided by Fisher information
to measure sparsity.
Beyond static masks,
Fish-DIP \cite{das2023fishdip}
further allows the Fisher information-based mask
to be updated dynamically during training.
Inspired by the lottery ticket hypothesis
\cite{frankle2018lottery},
LF-SFT \cite{ansell2021composable}
finds that sparse masks
obtained by selecting the parameters
with the largest changes
\emph{after} fine-tuning on a task
can be transferred to other tasks.
However,
this approach requires full fine-tuning
on an initial task,
which may not be feasible
for resource-constrained settings.
This paper explores the design principles
for constructing the sparsity mask
with \emph{low-cost} salience metrics
and the impact of dynamic versus static masks
on the fine-tuning process.

Finally,
sparsity-based adapters
also allow highly granular control over trainable parameters,
and can enable the use of existing knowledge transfer techniques,
such as mixtures of sparse experts \cite{xu2024mome}
and multi-task learning with sparse masks \cite{sun2020sparse}
in LLMs.

\subsection{Salience Proxies for Sparsity Masking}
The extensive research on low-cost salience metrics
for fine-grained network pruning
has provided a rich set of pruning-at-initialization metrics
to determine the importance of neural network parameters.
These metrics can be broadly classified
into first- and second-order categories.
First-order metrics
include weight magnitude \cite{han2015deep},
connection sensitivity (SNIP) \cite{lee2018snip},
foresight connection sensitivity (FORCE) \cite{jorge2021force},
Taylor-FO \cite{molchanov2019taylorfo},
SynFlow \cite{tanaka2020synflow},
and finally,
the gradient of the loss with respect to the weight.
Second-order metrics
comprise GRaSP \cite{wang2020grasp}
and Fisher information-based metrics \cite{liu2021group}.
Coincidentally,
both FishMASK \cite{sung2021fishmask}
and Fish-DIP \cite{das2023fishdip}
propose to use Fisher information
to construct the sparsity mask:
while FishMASK uses a static mask,
Fish-DIP further allows the mask
to be updated periodically during fine-tuning.
These metrics are designed
to identify important parameters or connections
in a neural network.
In this paper,
we explore the impact of these salience metrics
on fine-tuning
by using them to construct sparse masks
for PEFT\@.

\section{Method}\label{sec:method}

\subsection{Problem Formulation}\label{sec:method:problem}

Given a pretrained model \( f_{\weight_0} \)
with initial parameters \( \weight_0 \),
a dataset \( \trainset \),
and a downstream task loss function \( \loss \),
the goal of \emph{sparse} parameter-efficient fine-tuning (SPEFT)
is to find a set of sparse trainable parameters
\( \sparseweight \),
that minimizes the loss function
on the training dataset \( \trainset \):
\begin{equation}
    \sparseweight^\star = {\argmin}_{\sparseweight}
        \mathbb{E}_{\parens{\x, y} \sim \trainset} \bracks*{
            \loss\parens{f_{\weight_0 + \sparseweight}(\x); y}
        }.
\end{equation}
To ensure the sparsity of \( \sparseweight \),
we constrain it
with \( \indicator\bracks{\sparseweight \neq 0} = \mask \),
where \( \indicator\bracks{\cdot} \) is the indicator function,
\( \mask \in \braces{0, 1}^{d_1 \times d_2} \)
is the sparsity mask
with \( \verts{\mask}_0 \leq \density \ll d_1 \times d_2 \),
where \( \density \) is the number of non-zero entries.
This opens up the flexibility of \( \mask \) design,
\ie{}, selecting the non-zero locations in \( \sparseweight \)
to update during fine-tuning,
which can be determined by various salience metrics
as discussed below in \Cref{sec:method:salience}.

\subsection{%
    Salience Metrics
}\label{sec:method:salience}

In this section,
we describe the \nummetrics{} salience metrics
which can be used
to determine the importance of weights \( \weight \).
Assume that \( \x \) is the sampled input,
\( \ell \triangleq \loss\parens{f_\weight(\x); y} \)
is the loss function,
\( \odot \) denotes element-wise multiplication,
and \( \verts{\cdot} \)
denotes the element-wise absolute value.
For simplicity,
we also assume all data-aware metrics
to be expectations over the training dataset
\( \parens{\x, y} \sim \trainset \),
which can be approximated by sampling from it.
We have the following
\numfirstorder{}
\first{}-order salience metrics:
\begin{itemize}
    \item Magnitude:
    \( \verts\weight \),
    where simply the magnitude
    (\ie{}, absolute value)
    of the weight is used.

    \item Gradient:
    \( \diff{\ell}{\weight} \),
    which is the gradient of the loss
    with respect to the weight \( \weight \).

    \item SNIP (single-shot network pruning):
    \( \verts*{\diff{\ell}{\weight} \odot \weight} \),
    the connection sensitivity metric
    proposed in \cite{lee2018snip}
    to determine the importance of weights.

    \item FORCE (foresight connection sensitivity)\@:
    \( -\diff{\ell}{\weight} \odot \weight \),
    introduced in \cite{jorge2021force}.

    \item Taylor-FO (Taylor first-order expansion):
    \( \parens*{\diff{\ell}{\weight} \odot \weight}^2 \),
    derived from the \first{}-order Taylor expansion
    of the loss \cite{molchanov2019taylorfo}.

    \item SynFlow (iterative synaptic flow pruning):
    \(
        \diff{}{\weight}\bracks*{
            \mathbf{1}^\top
            \parens*{
                \Pi_{l=1}^L \verts{\weight^{(l)}}
            }
            \mathbf{1}
        }
        \odot \weight
    \),
    where \( \weight^{(l)} \) denotes the weights
    of the \ordinal{\( l \)} layer,
    and \( L \) denotes the number of layers.
    A data-free metric proposed in \cite{tanaka2020synflow}
    to model synaptic flow.
\end{itemize}

In addition,
the \second{}-order salience metrics
are computed as follows,
where
\(
    H \triangleq
    \frac{
        \partial^2 \! \loss\parens{f_\weight(\x); y}
    }{
        \partial\weight
        \partial\weight^\top
    }
\)
denotes the Hessian matrix:
\begin{itemize}
    \item GRaSP (gradient signal preservation):
    \( - \parens*{H \diff{\ell}{\weight}} \odot \weight \),
    which is a \second{}-order metric
    proposed in \cite{wang2020grasp}
    that aims to preserve gradient signals
    rather than the loss value.

    \item Fisher information:
    \( \parens*{\diff{\ell}{\weight}}^2 \),
    which uses the Fisher information
    to determine the importance of weights
    \cite{sung2021fishmask,das2023fishdip}.
\end{itemize}

\subsection{Sparsity Masking}\label{sec:method:masking}

\textbf{Global Sparsity Masking}
Given a salience metric \( \smetric(\weight) \)
of the weight \( \weight \) defined in \Cref{sec:method:salience},
we can construct the sparse binary mask \( \mask \)
by selecting the top \( \density \in (0, 1] \) fraction
of the salience metric values,
\ie{}, \( \density \) denotes the density level,
namely:
\begin{equation}
    \mask = \indicator\bracks*{
        \sval \geq \topk_\density\parens{\sval}
    },
    \text{where}\,
    \sval = \smetric(\weight).
\end{equation}
Here \( \indicator \) is the indicator function,
and \( \topk_\density \) selects the top \( \density \) values.

\textbf{Local Sparsity Masking}
Instead of ranking the salience metric values
across all weight values,
alternatively,
we can construct layer-wise masks \( \mask^{(l)} \)
for the individual weights \( \weight^{(l)} \)
in each layer \( l \),
where each layer has a shared sparsity \( \density \),
and the top \( \density \) values are selected
from the salience metric values
of the weights in that layer:
\begin{equation}
    \mask^{(l)} = \indicator\bracks*{
        \sval^{(l)} \geq
            \topk_\density\parens{\sval^{(l)}}
    },
    \text{where}\,
    \sval^{(l)} = \smetric(\weight^{(l)}).
\end{equation}
Here,
\( \weight \) is decomposed into layer-wise weights
\( \bracks*{\weight^{(1)}, \ldots, \weight^{(L)}} \)
and \( \mask^{(l)} \) and \( \weight^{(l)} \)
respectively denotes the mask and weights
of the \ordinal{\( l \)} layer.

\subsection{%
    Static \versus{} Dynamic Masks
}\label{sec:method:static-or-dynamic}

Beyond generating a static mask
using the above approach prior to fine-tuning,
which remains fixed throughout the training process,
we can also explore the use of dynamic masks,
which are updated periodically during training.
The dynamic mask
can be refreshed at specific intervals
by the following procedure:
first,
we apply the current trained weights
to the model;
we then re-rank the salience metric values
with these weights,
the top \( \density \) values
are then selected to form a new mask
using the updated salience metric values;
subsequently,
the fine-tuning process continues
with the new mask.
Notably,
after updating the dynamic masks,
we also need to reinitialize memory-based optimizers
in order to avoid applying incorrect momentum values
to the newly adapted sparse weights.

\subsection{The SPEFT Algorithm}\label{sec:method:algorithm}

\begin{algorithm*}[t]
    \caption{%
        Sparse Parameter-Efficient Fine-Tuning (SPEFT)
    }\label{alg:overview}
    \begin{algorithmic}[1]
        \Require{%
            Pretrained model \( f_{\weight_0} \),
            training dataset \( \trainset \),
            batch size \( B \),
            loss function \( \loss \),
            salience metric \( \smetric \),
            sparsity level \( \density \),
            fine-tuning steps \( T \),
            fine-tuning learning rate \( \lr \),
            mask update interval \( I \)
        }
        \State{\(
            \sparseweight \gets \mathbf{0};
            \weight \gets \weight_0
        \)}
        \algorithmiccomment{Initialize weights}
        \For{\( t = 1 \) to \( T \)}
            \algorithmiccomment{For each fine-tuning step\ldots}
            \If{\( t = 1 \vee \parens{I \geq 0 \wedge \parens{t\!\mod I = 0}} \)}
                \algorithmiccomment{%
                    \textbf{If salience masks should update\ldots}
                    (\Cref{sec:method:static-or-dynamic})}
                \State{\(
                    \parens{\weight, \sparseweight} \gets
                        \parens{\weight + \sparseweight, \mathbf{0}}
                \)}
                \algorithmiccomment{Apply sparse weights to model}
                \State{\(
                    \sval \gets \smetric(\weight)
                \)}
                \algorithmiccomment{%
                    \textbf{Compute salience values for all weights}
                    (\Cref{sec:method:salience})
                }
                \State{\(
                    \mask \gets \indicator\bracks*{
                        \sval \geq \topk_{\density}\parens{\sval}
                    }
                \)}
                \algorithmiccomment{%
                    \textbf{Update mask by top-\( \density \) values}
                    (\Cref{sec:method:masking})
                }
            \EndIf
            \State{\( \parens{\x_{[1:B]}, y_{[1:B]}} \gets \minibatch(\trainset) \)}
            \algorithmiccomment{Sample mini-batch}
            \State{\(
                \ell \gets \frac1B \sum_{b=1}^B \loss\parens{
                    f_{\weight + \sparseweight}(\x_b); y_b
                }
            \)}
            \algorithmiccomment{Forward pass}
            \State{\(
                \sparseweight \gets
                    \mathrm{Opt}\parens*{
                        \lr,
                        \sparseweight,
                        \mask \odot \diff{\ell}{\sparseweight}
                    }
            \)}
            \algorithmiccomment{%
                \textbf{Parameter-efficient optimization of sparse weights}}
        \EndFor
        \algorithmiccomment{%
            \textbf{NOTE}:
            only need to compute non-zero entries of \( \mask \)
            for the gradient}
        \State{\Return{\( \weight + \sparseweight \)}}
        \algorithmiccomment{Return fine-tuned model}
    \end{algorithmic}
\end{algorithm*}

\Cref{alg:overview}
provides an overview of the proposed SPEFT algorithm
to fine-tune models with sparse weight adaptations.
The algorithm takes as input
a pretrained model \( f_{\weight_0} \),
an optimizer \( \mathrm{Opt} \),
a training dataset \( \trainset \),
a batch size \( B \),
a loss function \( \loss \),
a salience metric \( \smetric \),
a sparsity level \( \density \),
the number of fine-tuning steps \( T \),
the learning rate \( \lr \),
and the mask update interval \( I \).
The algorithm
begins by initializing the sparse weights
\( \sparseweight \) to zero (line 1),
and then iterates for \( T \) steps (line 2).
In each iteration,
the algorithm first checks
if it is the initial iteration,
which requires updating the mask,
or if it is at the correct interval
for iterative dynamic mask updates (line 3).
If either of these conditions is true,
the algorithm applies the current sparse weights
to the model (line 4),
evaluates the new salience values \( \sval \) (line 5),
and updates the salience mask \( \mask \)
for the updated weights,
on the sparsity level \( \density \) (line 6).
After updating the mask,
the training step
follows by sampling a mini-batch \( \braces{\x, y} \)
from the training dataset (line 8),
and learns the sparse weights \( \sparseweight \) (line 9)
using the optimizer \( \mathrm{Opt} \)
(\eg{}, stochastic gradient descent, Adam, \etc{}).
Here,
\( \mask \odot \lr \diff{\ell}{\sparseweight} \)
where \( \odot \) denotes element-wise multiplication.
In terms of actual implementation,
only the non-zero entries
in \( \diff{\ell}{\sparseweight} \)
dictated by the mask \( \mask \)
are computed and updated.
Finally,
the algorithm returns the fine-tuned model
\( \weight_0 + \sparseweight \).


\section{Experimental Results}\label{sec:results}


\paragraph{Models}
We evaluated our approaches and baselines
over a set of models,
including fine-tuned OPT variants
(-125m, -350m, and -1.3b) \cite{zhang2022opt},
BERT-base-uncased \cite{devlin2019bert}
and RoBERTa-base \cite{liu2019roberta},
for the GLUE \cite{wang2019glue} benchmark,
and fine-tuned Gemma2-2b \cite{team2024gemma}
and Qwen2-7b \cite{yang2024qwen2},
to evaluate on the Massive Multitask Language Understanding (MMLU) benchmark
\cite{hendrycks2021mmlu}
and GSM8K \cite{cobbe2021gsm8k},
a dataset of grade school math problems.
Moreover,
we fine-tuned Llama3-8b \cite{grattafiori2024llama3}
to evaluate on the HumanEval \cite{chen2021humaneval} 
and MBPP \cite{austin2021mbpp} benchmarks.
In addition to sparse PEFT methods
presented in this paper,
we further include LoRA \cite{hu2021lora}
and PiSSA \cite{meng2024pissa}
as low-rank adapter baselines
for comparison.

\paragraph{Benchmarks}
To show the generality of our approach,
we chose GLUE, MMLU, GSM8K, HumanEval and MBPP as benchmarks for evaluation.
For the GLUE \cite{wang2019glue} benchmark,
six representative tasks
with large sizes are selected:
single-sentence task SST-2,
inference tasks QNLI, MNLI,
similarity and paraphrase tasks
MRPC, STS-B and QQP%
\footnote{We did not evaluate CoLA and RTE
because these datasets are too small
and require special treatments
such as fine-tuning RTE
using an MNLI checkpoint~\cite{lan2019albert}.}.
For the MMLU \cite{hendrycks2021mmlu} benchmark,
it contains questions covering 57 subjects
across STEM, the humanities, the social sciences, and others.
It is designed to test the model's ability
to handle various types of language data and complex problems.
We fine-tuned Gemma-2-2b, Qwen2-7b
on either the Alpaca \cite{alpaca}
or OASST2 \cite{köpf2023oasst} conversational datasets,
and then evaluated them on all tasks in MMLU\@.
We fine-tuned Gemma2-2b
on the MetaMathQA \cite{yu2024metamath} dataset
and evaluated on GSM8K (5-shots)
to assess the models' multi-step mathematical reasoning ability.
Furthermore,
we fine-tuned Llama3-8b on the CodeFeedback \cite{zheng2024codefeedback},
to evaluate on the HumanEval and MBPP 
which are aimed to test the code generation ability of models.
In the results,
we reported the match accuracy for MNLI,
Pearson correlation for STS-B,
flexible extract and strict match scores for GSM8K,
Pass@1 for HumanEval and MBPP,
and accuracy values for other tasks.

\paragraph{Baselines}
We chose LoRA \cite{hu2021lora}
and PiSSA \cite{meng2024pissa}
as the competing low-rank baselines
across models and benchmarks.
By default in all comparisons,
SPEFT methods use global sparsity ranking with static masks.
For statistical significance,
we repeated each experiment \numtrials{} times
for OPT-\{125m,350m\}, BERT-base-uncased, and RoBERTa-base,
and reported average metrics and their standard deviations.

\paragraph{Ablation Analyses}
We used the most reliable salience metric,
\ie{}, gradient-based,
in further experiments to explore questions
related to dynamic \versus{} static masks,
and global \versus{} local sparsity
in \Cref{sec:results:ablation}.
Additionally,
we also explored the efficiency-performance trade-off 
between LoRA, PiSSA and sparse baselines in \Cref{app:ablation}.

\paragraph{Hyperparameters}
Our SPEFT methods
introduce a hyperparameter \( \density \),
the percentage of trainable parameters.
To ensure a fair comparison,
we fixed \( \density \) of our SPEFT methods
to use the same amounts of trainable parameters
as LoRA and PiSSA on every model,
and kept the remaining hyperparameters always the same.
For example,
for the RoBERTa-base model,
we performed a grid sweep over learning rates
from \( 5 \times 10^{-4} \) to \( 5 \times 10^{-5} \)
to search for the best.
Details about the hyperparameter settings
can be found in \Cref{app:setup}.

\subsection{Main Results}\label{sec:results:main}

Our experiments results
on OPT-350m and BERT-base-uncased
can be seen in \Cref{tab:glue_opt350m_bert}.
For additional results
on RoBERTa-base, OPT-125m and OPT-1.3b,
please refer to \Cref{%
    tab:glue_roberta,%
    tab:glue_opt125m,%
    tab:glue_opt1.3b}
in \Cref{app:results}.
Across all models,
we observed that among all the approaches,
gradient-based SPEFT
has the best average accuracy,
higher than LoRA and PiSSA.
For instance,
in OPT-125m and OPT-350m,
gradient-based SPEFT achieves \( 86.92\% \) and \( 88.45\% \),
that are higher than the best competing SPEFT methods
by \( 0.73\% \) and \( 0.85\% \) respectively.
Particularly on OPT-350m,
gradient-based SPEFT has the best performance
on MNLI, MRPC, SST-2, and STS-B,
On QNLI and QQP,
LoRA has the best performance
while gradient-based SPEFT
has a good performance close to it.
This shows that although LoRA
shows excellent performance on certain tasks,
SPEFT methods,
particularly with the gradient salience metric,
could further push the limit,
achieving better results in accuracy.
On BERT-base-uncased,
we found that while SPEFT with Fisher-Info salience metric
outperforms gradient-based SPEFT on QNLI, QQP and SST-2,
it has a large gap in performance in the remaining tasks,
making gradient-based SPEFT a more reliable and desirable choice.
Similar results are also observed
for other OPT variants in \Cref{tab:glue_opt125m,tab:glue_opt1.3b}
and RoBERTa-base in \Cref{tab:glue_roberta}
of \Cref{app:results}.
\unskip\begin{table*}[t]
\centering
\cvspace{-1em}
\adjustbox{max width=\linewidth}{%
\begin{tabular}{l|cccccc|cc}
    \toprule
    \textbf{Method}
        & \textbf{MNLI}   & \textbf{MRPC}   & \textbf{QNLI}
        & \textbf{QQP}    & \textbf{SST-2}  & \textbf{STS-B}
        & \textbf{Avg.}   & \textbf{\#} \\
    \midrule
    \midrule
    \multicolumn{8}{c}{\textbf{OPT-350m} (Trainable = 0.35\%)} \\
    \midrule
    LoRA
        & \std{83.56}{.07}
        & \std{84.56}{.49}
        & \beststd{89.69}{.11}
        & \beststd{89.66}{.04}
        & \std{93.87}{.06}
        & \std{88.57}{.99}
        & \std{88.32}{.29} & 2 \\
    PiSSA
        & \std{83.45}{.06}
        & \std{83.09}{.52}
        & \std{89.38}{.06}
        & \beststd{89.66}{.02}
        & \std{93.58}{.09}
        & \std{88.39}{.52}
        & \std{87.93}{.21} & 1 \\
    \midrule
    Magnitude
        & \std{79.34}{.41}
        & \std{71.57}{.13}
        & \std{86.45}{.06}
        & \std{87.68}{.01}
        & \std{91.98}{.12}
        & \std{45.04}{3.39}
        & \std{77.01}{.51} & 0 \\
    Gradient
        & \beststd{83.86}{.06}
        & \beststd{84.80}{.55}
        & \std{89.68}{.01}
        & \std{89.51}{.01}
        & \std{93.93}{.12}
        & \beststd{88.95}{.25}
        & \beststd{88.45}{.02} & \best{3} \\
    SynFlow
        & \std{77.45}{.05}
        & \std{77.94}{.49}
        & \std{83.19}{.03}
        & \std{88.03}{.02}
        & \std{92.32}{.18}
        & \std{79.18}{.63}
        & \std{83.02}{.22} & 0 \\
    SNIP
        & \std{83.40}{.05}
        & \std{83.09}{.37}
        & \std{89.68}{.22}
        & \std{89.37}{.02}
        & \std{93.75}{.06}
        & \std{86.32}{.04}
        & \std{87.60}{.10} & 0 \\
    FORCE
        & \std{83.25}{.08}
        & \std{82.60}{.62}
        & \std{89.75}{.30}
        & \std{89.50}{.03}
        & \beststd{94.04}{.69}
        & \std{85.53}{.18}
        & \std{87.44}{.26} & 0 \\
    Taylor-FO
        & \std{83.31}{.08}
        & \std{83.09}{.37}
        & \std{89.68}{.22}
        & \std{89.37}{.02}
        & \std{93.75}{.06}
        & \std{86.32}{.04}
        & \std{87.59}{.12} & 0 \\
    \midrule
    GRaSP
        & \std{74.78}{.27}
        & \std{83.58}{.49}
        & \std{84.46}{.39}
        & \std{89.38}{.03}
        & \beststd{94.04}{.01}
        & \std{86.97}{.01}
        & \std{85.54}{.20} & 1 \\
    Fisher-Info
        & \std{35.45}{1.35}
        & \std{84.31}{.61}
        & \std{88.12}{.34}
        & \std{86.34}{.41}
        & \std{87.16}{.35}
        & \std{88.61}{.02}
        & \std{78.33}{.51} & 0 \\
    \midrule
    \midrule
    \multicolumn{9}{c}{\textbf{BERT-base-uncased} (Trainable = 0.27\%)} \\
    \midrule
    LoRA
        & \beststd{81.45}{.41}
        & \std{88.48}{1.03}
        & \std{89.57}{.35}
        & \std{87.77}{.54}
        & \std{91.82}{.14}
        & \std{84.07}{1.11}
        & \std{87.19}{.30} & 1 \\
    PiSSA
        & \std{81.08}{.27}
        & \std{87.75}{.43}
        & \std{90.19}{.30}
        & \std{88.14}{.33}
        & \std{91.51}{.08}
        & \beststd{85.12}{.26}
        & \std{87.30}{.18} & 1 \\
    \midrule
    Magnitude
        & \std{77.09}{.24}
        & \std{68.88}{.25}
        & \std{86.60}{.07}
        & \std{85.56}{.50}
        & \std{90.14}{.02}
        & \std{37.59}{1.93}
        & \std{74.31}{.33} & 0 \\
    Gradient
        & \std{80.99}{.12}
        & \beststd{89.46}{.48}
        & \std{89.90}{.26}
        & \std{87.48}{.13}
        & \std{91.63}{.01}
        & \std{85.08}{.06}
        & \beststd{87.42}{.15} & 2 \\
    SynFlow
        & \std{70.85}{.21}
        & \std{71.33}{.25}
        & \std{83.49}{.04}
        & \std{83.69}{.16}
        & \std{90.08}{.29}
        & \std{74.55}{.36}
        & \std{79.00}{.12} & 0 \\
    SNIP
        & \std{80.74}{.20}
        & \std{79.90}{1.47}
        & \std{89.39}{.08}
        & \std{87.27}{.25}
        & \std{91.57}{.06}
        & \std{80.92}{.41}
        & \std{84.96}{.18} & 0 \\
    FORCE
        & \std{80.25}{.09}
        & \std{78.31}{.86}
        & \std{88.98}{.15}
        & \std{87.04}{.38}
        & \std{91.57}{.17}
        & \std{79.21}{.24}
        & \std{84.23}{.15} & 0 \\
    Taylor-FO
        & \std{80.74}{.20}
        & \std{79.90}{1.47}
        & \std{89.39}{.08}
        & \std{87.27}{.25}
        & \std{91.57}{.06}
        & \std{80.87}{.46}
        & \std{84.96}{.18} & 0 \\
    \midrule
    GRaSP
        & \std{79.37}{.27}
        & \std{77.95}{1.72}
        & \std{87.50}{1.12}
        & \std{87.03}{.41}
        & \std{91.35}{.52}
        & \std{79.67}{1.43}
        & \std{83.81}{.59} & 0 \\
    Fisher-Info
        & \std{79.83}{.16}
        & \std{87.75}{.74}
        & \beststd{90.46}{.22}
        & \beststd{88.78}{.25}
        & \beststd{91.86}{.34}
        & \std{82.79}{.63}
        & \std{86.91}{.18} & \best{3} \\
    \bottomrule
\end{tabular}}
\cvspace{-1em}
\caption{%
    Comparing the salience metrics
    on OPT-350m (with 0.35\% trainable parameters)
    and BERT-base-uncased (with 0.27\% trainable parameters)
    for various GLUE tasks.
    For reference,
    we provide the LoRA and PiSSA baselines
    with the same number of trainable parameters
    for each model.
    The ``\#'' column
    denotes the number of best performing tasks
    for each method.
    The best result of each column
    is highlighted in bold.
    ``Avg.'' reports the average score across all tasks,
    and their average standard deviations.
}\label{tab:glue_opt350m_bert}
\end{table*}

Notably,
for both causal and masked language models,
\textbf{sparsity-based PEFT can outperform low-rank adapters},
and the \textbf{gradient-based SPEFT
shows the strongest performance}
compared to other methods,
closely followed by LoRA and PiSSA,
which is consistent across all models.
In addition,
the gradient-based SPEFT
outperformed LoRA and PiSSA in several tasks,
highlighting its effectiveness
across different model sizes.
The comprehensive results table
for these models and tasks
underlines the consistent performance edge
of gradient-based SPEFT,
making it a reliable choice
for a wide range of NLP tasks.

\subsection{Larger Scale Models}\label{sec:results:larger_scale}

For larger models,
we evaluated all methods on Gemma2-2b and Qwen2-7b,
and show the results in \Cref{tab:mmlu}.
The results indicate that larger models
can also benefit from SPEFT
with the gradient-based saliency method,
which outperforms other sparse training methods and LoRA.
\begin{table}[ht]
\centering
\cvspace{-.5em}
\adjustbox{max width=\linewidth}{%
\begin{tabular}{l|cc|cc|c}
    \toprule
    \textbf{Model}
        & \multicolumn{2}{c|}{\textbf{Gemma2-2b}}
        & \multicolumn{2}{c|}{\textbf{Qwen2-7b}}
        & \multirow{2}{*}{\textbf{Avg.}} \\
    \textbf{Dataset}
        & \textbf{Alpaca} & \textbf{OASST2}
        & \textbf{Alpaca} & \textbf{OASST2} \\
    \midrule
    \midrule
    LoRA
        & 53.07 & 52.59
        & 69.77 & 70.42
        & 61.46 \\
    \midrule
    Gradient
        & \best{53.11} & \best{53.11}
        & \best{70.96} & 70.55
        & \best{61.93} \\
    SynFlow
        & 52.84 & 53.07
        & 69.80 & 70.66
        & 61.59 \\
    Magnitude
        & 52.97 & 53.03
        & 70.12 & \best{70.76}
        & 61.72 \\
    SNIP
        & 52.81 & 52.89
        & 68.75 & 70.52
        & 61.24 \\
    FORCE
        & 52.79 & 52.88
        & 69.01 & 70.53
        & 61.30 \\
    Taylor-FO
        & 52.81 & 52.96
        & 68.75 & 69.10
        & 60.91 \\
    \midrule
    GRaSP
        & 52.38 & 52.60
        & 66.69 & 69.91
        & 60.40 \\
    Fisher-Info
        & 52.70 & 52.65
        & 66.45 & 69.10
        & 60.23 \\
    \bottomrule
\end{tabular}}
\cvspace{-1em}
\caption{%
    Comparing the salience metrics
    on Gemma2-2b and Qwen2-7b
    respectively with 0.97\% and 0.53\% trainable parameters.
    We fine-tuned models
    on either Alpaca or OASST2
    and evaluated on 5-shot MMLU.
    For reference,
    we provide the LoRA baselines
    with the same number of trainable parameters
    for each combination.
}\label{tab:mmlu}
\end{table}

To evaluate on the text generation task,
We fine-tuned Gemma2-2b
with our methods on MetaMathQA
and evaluated on 5-shot GSM8K.
We also provide the results
of the pretrained model (without fine-tuning)
and LoRA as baselines.
The results are shown in \Cref{tab:gsm8k}.
It can be seen that the sparse adapters
outperformed the LoRA baseline,
with the gradient-based SPEFT method
leading the pack with the best performance.
Furthermore,
for code generation tasks,
we fine-tuned Llama3-8b with our methods
and evaluated on HumanEval and MBPP benchmarks.
The results are shown in \Cref{tab:code_generation} of \Cref{app:results}.
Notably,
the \textbf{lead by sparse adapters widens
as the task complexity increases},
which demands token sequence generation
with multi-step reasoning.
\begin{table}[!ht]
\centering
\cvspace{-.5em}
\adjustbox{max width=\linewidth}{%
\begin{tabular}{l|cc|cc|c}
    \toprule
    \textbf{Method}
        & \textbf{Flexible Extract}
        & \textbf{Strict Match}
        & \textbf{Avg.} \\
    \midrule
    \midrule
    Pretrained
        & 24.56 & 17.66
        & 21.11 \\
    LoRA
        & 39.20 & 28.81
        & 34.00 \\
    \midrule
    Gradient
        & \best{50.27} & \best{37.15}
        & \best{43.71} \\
    SynFlow
        & 37.76 & 27.75
        & 32.75 \\
    Magnitude
        & 37.45 & 27.07
        & 32.26 \\
    SNIP
        & 39.80 & 29.64
        & 34.72 \\
    FORCE
        & 39.88 & 29.95
        & 34.91 \\
    Taylor-FO
        & 40.33 & 30.25
        & 35.29 \\
    \midrule
    GRaSP
        & 50.15 & 37.03
        & 43.59 \\
    Fisher-Info
        & 41.47 & 30.25
        & 35.86 \\
    \bottomrule
\end{tabular}}
\cvspace{-1em}
\caption{%
    Comparing the salience metrics
    on Gemma2-2b with 0.97\%
    trainable parameters.
    We fine-tuned the model
    on MetaMathQA
    and evaluated on 5-shot GSM8K.
    For reference,
    we provide pretrained model (without fine-tuning)
    and the LoRA baseline
    with the same number of trainable parameters.
}\label{tab:gsm8k}
\end{table}

\subsection{%
    Exploration of masking strategies
}\label{sec:results:ablation}

\begin{table}[!ht]
\centering
\cvspace{-.5em}
\adjustbox{max width=\linewidth}{%
\begin{tabular}{c|cccccc|c}
    \toprule
        & \textbf{MNLI} & \textbf{MRPC}
        & \textbf{QNLI} & \textbf{QQP}
        & \textbf{SST-2} & \textbf{STS-B}
        & \textbf{Avg.} \\
    \midrule
    \midrule
    \multicolumn{8}{c}{\textbf{OPT-125m} (Trainable = 0.35\%)} \\
    \midrule
    \textbf{SG}
        & \best{81.41} & \best{83.82} & 88.58 & 88.71 & \best{91.44} & 87.55 & \textbf{86.92} \\
    \textbf{SL}
        & \best{81.41} & 81.86 & \best{88.94} & \best{88.76} & 91.40 & 87.38 & 86.63 \\
    \textbf{DG}
        & 77.71 & 82.84 & 83.80 & 87.36 & 89.33 & \best{88.28} & 84.89 \\
    \textbf{DL}
        & 69.26 & 73.53 & 80.56 & 84.82 & 86.35 & 87.15 & 80.28 \\
    \midrule
    \midrule
    \multicolumn{8}{c}{\textbf{OPT-350m} (Trainable = 0.35\%)} \\
    \midrule
    \textbf{SG}
        & 83.86 & 84.80 & \best{89.68} & 89.51 & 93.93 & \best{88.95} & 88.46 \\
    \textbf{SL}
        & \best{84.31} & 83.33 & 90.63 & \best{90.97} & \best{94.50} & 88.52 & \textbf{88.71} \\
    \textbf{DG}
    & 78.03 & 85.29 & 89.22 & 84.24 & 91.51 & 88.54 & 86.14 \\
    \textbf{DL}
    & 78.86 & 71.57 & 80.84 & 84.52 & 87.27 & 87.52 & 81.76 \\
    \midrule
    \midrule
    \multicolumn{8}{c}{\textbf{BERT-base-uncased} (Trainable = 0.27\%)} \\
    \midrule
    \textbf{SG}
    & 80.99 & \best{89.46} & 89.90 & 87.48 & 91.63 & 85.08 & 87.42 \\
    \textbf{SL}
    & 74.58 & 85.54 & 89.62 & 83.41 & 91.06 & 85.79 & 85.00 \\
    \textbf{DG}
    & \best{83.17} & \best{89.46} & \best{90.32} & \best{90.27} & \best{92.20} & 84.20 & \textbf{88.27} \\
    \textbf{DL}
    & 72.80 & 86.52 & 83.49 & 82.51 & 90.25 & \best{85.95} & 83.59 \\
    \bottomrule
\end{tabular}}
\cvspace{-1em}
\caption{%
    Results of OPT-125m, OPT-350m and BERT-base-uncased
    with fixed or dynamic gradient masks
    and global or local sparsity
    on various GLUE tasks.
    The dynamic strategy
    will update the gradient mask every 1000 train steps.
    ``S / D'': static / dynamic masks,
    ``G / L'': global / local sparsity.
    Runs were repeated \numtrials{} times
    and all results have a standard deviation of $< 0.5\%$.
}\label{tab:mask_strategies}
\end{table}

Based on the comparisons with SPEFT in \Cref{sec:results:main},
which showed that gradient-based SPEFT is the best-performing method,
we would use it for ablation studies
of dynamic \versus{} static masks,
and global \versus{} local sparsity.
In this section,
we delve into the comparisons
between global and local sparsity
(\Cref{sec:method:masking})
and also static and dynamic masking strategies
(\Cref{sec:method:static-or-dynamic})
using gradient-based SPEFT,
the best-performing salience metric,
across OPT-125m, OPT-350m, and BERT-base-uncased.
Here,
we periodically update the masks
every \( I = 1000 \) steps
with 1024 training examples
to estimate the salience metrics.
The results are shown in \Cref{tab:mask_strategies}.

\paragraph{Dynamic \versus{} static masking}
The findings reveal that dynamic masking
offers only a slight performance advantage
in smaller models like BERT-base-uncased
but does not significantly outperform static masking
in larger models.
For instance,
on OPT-350m,
we actually see static masking
provides us a better averaged accuracy ($88.46$ and $88.71$)
compared to dynamic masking ($86.14$ and $81.76$).
Given that dynamic masking
requires more computational resources,
because of the periodic update
on sparsity masks,
the marginal performance gain
does not justify the extra cost,
especially for larger models.
Therefore,
static masking emerges
as a more practical and resource-efficient strategy,
providing substantial performance benefits
without the additional computational overhead.

\paragraph{Global \versus{} local sparsity}
With global sparsity,
SPEFT calculates the metrics
across all transformer layers,
ranks them collectively,
and makes only the highest-ranked ones trainable.
In the local approach,
metrics are sorted and ranked within each individual layer.
Our results
showed no significant difference in performance
between the two strategies.
For instance,
the results in BERT-base-uncased
suggests that global is superior,
by showing a better averaged accuracy
across the six GLUE tasks,
but the numbers in OPT-350m
suggest the reverse
under the static masking strategy.

\subsection{%
    Minimal Overhead for SPEFT
}\label{sec:results:overhead}

\paragraph{Computational overhead}
For all first-order salience metrics,
we use a few gradient evaluations
to compute the salience scores.
Specifically,
only 64 steps with a batch size of 16 per estimation
are needed
(1024 examples),
which is negligible compared
to the overall training cost.
For example,
this represents only 0.26\% and 0.97\% of the training time
for one epoch on MNLI and QNLI, respectively.
For static masks,
this computation is performed once before training;
for dynamic masking,
it is repeated once per \( I = 1000 \) steps.
Second-order metrics such as GRaSP and Fisher-Info
require \( 2\times \) the number of gradient evaluations
of first-order metrics
to compute the second-order gradients.
The magnitude metric requires no additional computation.
Finally,
we observed no statistically significant difference
in training time between the sparse methods and the LoRA baseline.

\paragraph{Memory overhead}
As we aligned the number of trainable parameters
across LoRA and the SPEFT methods,
the peak memory usage for both methods are mostly identical,
except that the SPEFT methods
require a small amount of additional memory overhead
to store the indices in CSR format.
In all experiments,
the overhead is less than 0.5\% of the peak memory usage.

\section{Discussion}\label{sec:discussion}

\paragraph{%
	The Trend of Supporting Sparse Computation
    as Hardware Intrinsics}
Numerous hardware vendors
have introduced specialized hardware features
with instruction set extensions
tailored for sparse matrix multiplication.
Especially in recently announced hardware devices.
Mainstream devices
like NVIDIA's A100 \cite{choquette2021nvidia},
H100 \cite{choquette2023nvidia},
and H200,
as well as offerings
from other major vendors
or emerging competitors
such as AMD's MI300 \cite{AMD}
and Cerebras' WSE2 \cite{selig2022cerebras},
are embracing this trend.
As hardware support for sparse computation advances,
the utility of sparsity-based PEFT,
or generally sparse training,
is poised to improve substantially.
This development
will enable both current and future strategies
to attain performance levels
closer to their full potential,
as these calculations won't require emulation
via dense computations,
allowing for closer realization
of theoretical speedups and savings on FLOPs.

\paragraph{The Role of Salience Measurements}
A fundamental element of this study
involves reevaluating certain design choices in SPEFT,
leading to the discovery
that straightforward designs,
such as first-order salience proxies,
emerge as the most effective methods.
Intriguingly,
selecting the most salient weights
in a neural network
has being a long-standing challenge,
one that dates back to early weight pruning research
by LeCun \etal{} in 1989 \cite{lecun1989optimal}.
It's notable
that the optimal saliency metric
seems to differ -- or arguably should differ --
among different task setups,
such as post-training weight pruning \cite{lecun1989optimal},
pruning at initialization \cite{snip, jorge2021force},
and zero-cost NAS proxies \cite{siems2020bench}.
The suggested practice
then should be to systematically
review a range of known and established proxies
to set a solid baseline
before designing a complex salience metric.

\section{Conclusion}\label{sec:conclusion}

We explored the efficacy
of various sparse parameter-efficient fine-tuning
(SPEFT) methods
in enhancing the performance of LLMs.
Our experiments
compared LoRA and PiSSA against SPEFT methods
with a range salience metrics,
and demonstrated that gradient-based SPEFT
consistently achieved superior accuracy
across different tasks and model architectures.
This demonstrates that,
although LoRA and PiSSA is effective in certain contexts,
SPEFT methods that leverage gradient information
can further optimize performance.
We also investigated the impact
of static versus dynamic sparsity masks,
concluding that while dynamic masks
do not significantly outperform static masks,
and they introduce additional training overhead.
Our findings suggest that static masks,
combined with the gradient-based salience metric,
provide a practical balance
between computational efficiency
and model accuracy.
Overall,
our research contributes
to the ongoing efforts in making model fine-tuning
more efficient and accessible,
particularly in resource-constrained settings.

\section{Acknowledgments}\label{sec:acknowledgments}

This work is supported in part
by the National Key R\&D Program of China (2023YFC3321600),
National Natural Science Foundation of China
(62376263, 62372443 and 62271496),
Guangdong Basic and Applied Basic Research Foundation
(2023B1515130002),
Natural Science Foundation of Guangdong
(2024A1515030209 and 2024A1515011970),
Shenzhen Science and Technology Innovation Commission
(JCYJ20230807140507015 and JCYJ20220531100804009).







\section{Limitations}\label{sec:limitations}

During the experiments,
we found that in a few training runs,
SPEFT seems less sensitive to hyperparameter changes
than LoRA,
\ie{},
on a range of hyperparameter sets,
SPEFT always improves model performance,
but LoRA fails.
Due to limited resources and time,
we did not run additional experiments
to explore this interesting observation.
We leave this exploration for future work.
Moreover,
similar investigations
on parameter efficient fine-tuning
could be conducted with non-language-based models
or other multimodal models,
such as vision large language models (VLLMs),
however,
these explorations
are beyond the current scope of this paper
and thus is left as future work.


\iftoggle{arxiv}{%
    \bibliographystyle{plainnat}
}{}
\bibliography{references}
\clearpage
\appendix
\section{Hyperparameters}\label{app:setup}

The hyperparameters we used
for all models are shown
in \Cref{tab:hparams:glue,tab:hparams:alpaca,tab:hyper_metamath,tab:hparams_code}.
Notably,
for all models,
the density $\density$ was set
to make sure the number of trainable parameters
across all methods was the same as the LoRA baseline.
\begin{table}[h]
\centering
\adjustbox{max width=\linewidth}{%
\begin{tabular}{ll|cc}
    \toprule
    Method & Dataset
        & \tlbox{MRPC \\ STS-B} & \tlbox{QNLI \\ SST-2 \\ MNLI \\ QQP} \\
    \midrule
    \multirow{7}{*}{Shared}
        & Optimizer & AdamW & AdamW \\
        & Warmup Ratio & 0 & 0 \\
        & LR Schedule & Linear & Linear \\
        & Batch Size & 16 & 64 \\
        & \# Epochs & 30 & 30 \\
        & Learning Rate & 4E-4 & 5E-5 \\
        & Max Seq. Len. & 512 & 196 \\
    \midrule
    \multirow{2}{*}{LoRA}
        & LoRA $r$ & 8 & 8 \\
        & LoRA $\alpha$ & 16 & 8 \\
    \midrule
    OPT-125m
        & Sparse $\density$ & 0.35\% & 0.35\% \\ 
    OPT-350m
        & Sparse $\density$ & 0.35\% & 0.35\% \\
    BERT-base
        & Sparse $\density$ & 0.27\% & 0.27\% \\
    RoBERTa-base
        & Sparse $\density$ & 0.24\% & 0.24\% \\
    \bottomrule
    \toprule
    \tlbox{Model \\ (Method)} & Hyperparameters & \multicolumn{2}{c}{All datasets} \\
    \midrule
    \multirow{6}{*}{Shared}
        & Optimizer & \multicolumn{2}{c}{AdamW} \\
        & Warmup Ratio & \multicolumn{2}{c}{0} \\
        & LR Schedule & \multicolumn{2}{c}{Linear} \\
        & Learning Rate & \multicolumn{2}{c}{5E-5} \\
        & \# Epochs & \multicolumn{2}{c}{30} \\
        & Batch Size & \multicolumn{2}{c}{16} \\
    \midrule
    OPT-1.3b
        & LoRA $ r $ & \multicolumn{2}{c}{8} \\
    (LoRA)
        & LoRA $ \alpha $ & \multicolumn{2}{c}{8} \\
    \midrule
    \tlbox{OPT-1.3b \\ (Sparse)}
        & Sparse $ \density $ & \multicolumn{2}{c}{0.18\%} \\
    \bottomrule
\end{tabular}}
\caption{%
    The hyperparameters we used
    for all models evaluated on the GLUE benchmark.
    The percentage of trainable parameters (\( \density \))
    for the sparse models
    are chosen to be the same as the LoRA models.
}\label{tab:hparams:glue}
\end{table}
\begin{table}[h]
\centering
\adjustbox{max width=\linewidth}{%
\begin{tabular}{ll|cc}
    \toprule
    \tlbox{Model \\ (Method)} & Hyperparameters & Alpaca & OASST2 \\
    \midrule
    \multirow{5}{*}{Shared}
        & Optimizer & \multicolumn{2}{c}{AdamW} \\
        & Warmup Ratio & \multicolumn{2}{c}{0.03} \\
        & LR Schedule & \multicolumn{2}{c}{Constant} \\
        & Batch Size & \multicolumn{2}{c}{16} \\
        & Max Seq. Len. & \multicolumn{2}{c}{1024} \\
    \midrule
    \multirow{4}{*}{\tlbox{Gemma2-2b \\ (LoRA)}}
        & \# Steps & \multicolumn{2}{c}{2000} \\
        & Learning Rate & \multicolumn{2}{c}{5E-5} \\
        & LoRA $ r $ & \multicolumn{2}{c}{64} \\
        & LoRA $ \alpha $ & \multicolumn{2}{c}{16} \\
    \midrule
    \multirow{3}{*}{\tlbox{Gemma2-2b \\ (Sparse)}}
        & \# Steps & \multicolumn{2}{c}{2000} \\
        & Learning Rate & 1E-5 & 5E-6 \\
        & Sparse $ \density $ & \multicolumn{2}{c}{0.97\%} \\ 
    \midrule
    \multirow{4}{*}{\tlbox{Qwen2-7b \\ (LoRA)}}
        & \# Epochs/Steps & 3 Epochs & 2000 Steps \\
        & Learning Rate & \multicolumn{2}{c}{5E-5} \\
        & LoRA $ r $ & \multicolumn{2}{c}{64} \\
        & LoRA $ \alpha $ & \multicolumn{2}{c}{16} \\
    \midrule
    \multirow{3}{*}{\tlbox{Qwen2-7b \\ (Sparse)}}
        & \# Epochs/Steps & 3 Epochs & 2000 Steps \\
        & Learning Rate & 5E-5 & 5E-6 \\
        & Sparse $ \density $ & \multicolumn{2}{c}{0.53\%} \\ 
    \bottomrule
\end{tabular}}
\caption{%
    The hyperparameters we used
    for Gemma2-2b and Qwen2-7b
    on Alpaca and OASST2.
The percentage of trainable parameters (\( \density \))
for the sparse models
are chosen to be the same as the LoRA models.
}\label{tab:hparams:alpaca}
\end{table}
\begin{table}[t]
    \centering
\adjustbox{max width=\linewidth}{%
\begin{tabular}{ll|c}
    \toprule
    \tlbox{Model \\ (Method)} & Hyperparameters & MetaMathQA \\
    \midrule
    & Optimizer & AdamW \\
    Shared & Warmup Ratio & 0.03 \\
    & LR Schedule & Linear \\
    \midrule
    \multirow{6}{*}{\tlbox{Gemma2-2b \\ (LoRA)}}
    & Batch Size & 16 \\
    & \# Epochs & 1 \\
    & Learning Rate & 2E-5 \\
    & LoRA $ r $ & 64 \\
    & LoRA $ \alpha $ & 16 \\
    & Max Seq. Len. & 1024 \\
    \midrule
    \multirow{5}{*}{\tlbox{Gemma2-2b \\ (Sparse)}}
    & Batch Size & 16 \\
    & \# Epochs & 1 \\
    & Learning Rate & 2E-5 \\
    & Sparse Top $ k $ & 0.18\% \\
    & Max Seq. Len. & 1024 \\
    \bottomrule
\end{tabular}}
\caption{%
    The hyperparameters we used for Gemma2-2b on MetaMathQA.
    The percentage of trainable parameters (\( \density \))
    for the sparse models
    are chosen to be the same as the LoRA models.
}\label{tab:hyper_metamath}
    \vspace{1em}
        \centering
    \adjustbox{max width=\linewidth}{%
    \begin{tabular}{ll|c}
        \toprule
        \tlbox{Model \\ (Method)} & Hyperparameters & CodeFeedback \\
        \midrule
        & Optimizer & AdamW \\
        Shared & Warmup Ratio & 0.03 \\
        & LR Schedule & Linear \\
        \midrule
        \multirow{6}{*}{\tlbox{Llama3-8b \\ (LoRA)}}
        & Batch Size & 16 \\
        & \# Epochs & 1 \\
        & Learning Rate & 2E-5 \\
        & LoRA $ r $ & 64 \\
        & LoRA $ \alpha $ & 16 \\
        & Max Seq. Len. & 512 \\
        \midrule
        \multirow{5}{*}{\tlbox{Llama3-8b \\ (Sparse)}}
        & Batch Size & 16 \\
        & \# Epochs & 1 \\
        & Learning Rate & 2E-5 \\
        & Sparse Top $ k $ & 0.67\% \\
        & Max Seq. Len. & 512 \\
        \bottomrule
    \end{tabular}}
    \caption{%
        The hyperparameters we used for Llama3-8b on CodeFeedback.
        The percentage of trainable parameters (\( \density \))
        for the sparse models
        are chosen to be the same as the LoRA models.
    }\label{tab:hparams_code}
\end{table}
\begin{table*}[t]
    \centering
    \adjustbox{max width=\linewidth}{%
    \begin{tabular}{l|cccccc|cc}
    \toprule
    \textbf{Method}
        & \textbf{MNLI}   & \textbf{MRPC}   & \textbf{QNLI}
        & \textbf{QQP}    & \textbf{SST-2}  & \textbf{STS-B}
        & \textbf{Avg.}   & \textbf{\#} \\
    \midrule
    \midrule
    LoRA
        & \std{86.52}{.06}
        & \std{89.46}{.73}
        & \std{92.11}{.29}
        & \std{88.70}{.15}
        & \std{93.81}{.23}
        & \std{90.30}{.01}
        & \std{90.15}{.25} & 0 \\
    PiSSA
        & \beststd{86.71}{.02}
        & \std{89.47}{.42}
        & \beststd{92.20}{.09}
        & \std{88.46}{.10}
        & \std{93.75}{.14}
        & \beststd{90.78}{.02}
        & \std{90.23}{.13} & \best{3} \\
    \midrule
    Magnitude
        & \std{82.58}{.46}
        & \std{31.62}{2.05}
        & \std{88.03}{.35}
        & \std{86.37}{.36}
        & \std{90.60}{.23}
        & \std{15.16}{2.64}
        & \std{65.73}{1.01} & 0 \\
    Gradient
        & \std{86.00}{.05}
        & \beststd{90.44}{.11}
        & \std{91.89}{.13}
        & \std{88.78}{.05}
        & \beststd{94.16}{.06}
        & \std{90.29}{.02}
        & \beststd{90.26}{.04} & 2 \\
    SynFlow
        & \std{75.53}{.02}
        & \std{70.34}{.12}
        & \std{84.37}{.01}
        & \std{85.19}{.02}
        & \std{91.80}{.29}
        & \std{76.92}{.44}
        & \std{80.69}{.17} & 0 \\
    SNIP
        & \std{85.97}{.01}
        & \std{87.01}{.25}
        & \std{91.34}{.01}
        & \std{88.31}{.06}
        & \std{93.92}{.29}
        & \std{87.52}{.16}
        & \std{89.01}{.08} & 0 \\
    FORCE
        & \std{85.64}{.05}
        & \std{85.29}{.37}
        & \std{91.31}{.04}
        & \std{88.39}{.04}
        & \std{93.75}{.06}
        & \std{86.52}{.15}
        & \std{88.48}{.07} & 0 \\
    Taylor-FO
        & \std{85.97}{.01}
        & \std{87.01}{.25}
        & \std{91.34}{.01}
        & \std{88.31}{.06}
        & \std{93.92}{.29}
        & \std{87.52}{.16}
        & \std{89.01}{.08} & 0 \\
    \midrule
    GRaSP
        & \std{79.07}{.02}
        & \std{84.80}{.25}
        & \std{87.88}{.02}
        & \std{88.45}{.12}
        & \std{93.52}{.06}
        & \std{86.81}{.24}
        & \std{86.76}{.04} & 0 \\
    Fisher-Info
        & \std{85.52}{.15}
        & \std{86.76}{.35}
        & \std{91.82}{.06}
        & \beststd{89.16}{.03}
        & \std{93.92}{.28}
        & \std{87.51}{.05}
        & \std{89.12}{.15} & 1 \\
    \bottomrule
    \end{tabular}}
    \caption{%
        Comparing the salience metrics
        on RoBERTa-base
        for various GLUE tasks
        with 0.24\% trainable parameters,
        following the same format as \Cref{tab:glue_opt350m_bert}.
    }\label{tab:glue_roberta}
\end{table*}
\begin{table*}[t]
    \centering
    \adjustbox{max width=\linewidth}{%
    \begin{tabular}{l|cccccc|cc}
    \toprule
    \textbf{Method}
        & \textbf{MNLI}   & \textbf{MRPC}   & \textbf{QNLI}
        & \textbf{QQP}    & \textbf{SST-2}  & \textbf{STS-B}
        & \textbf{Avg.}   & \textbf{\#} \\
    \midrule
    \midrule
    LoRA
        & \beststd{81.94}{.22}
        & \std{82.84}{.23}
        & \std{88.23}{.30}
        & \std{88.45}{.20}
        & \beststd{91.97}{.18}
        & \std{87.25}{.47}
        & \std{86.78}{.21} & 2 \\
    PiSSA
        & \std{81.56}{.11}
        & \std{83.33}{.30}
        & \std{87.99}{.32}
        & \std{88.17}{.15}
        & \beststd{91.97}{.11}
        & \std{86.87}{.39}
        & \std{86.65}{.15} & 1 \\
    \midrule
    Magnitude
        & \std{78.03}{3.14}
        & \std{76.35}{4.05}
        & \std{85.46}{1.62}
        & \std{86.56}{1.15}
        & \std{90.40}{.85}
        & \std{50.32}{2.42}
        & \std{77.85}{3.27} & 0 \\
    Gradient
        & \std{81.41}{.01}
        & \beststd{83.82}{.37}
        & \beststd{88.58}{.37}
        & \beststd{88.71}{.09}
        & \std{91.46}{.05}
        & \std{87.55}{.34}
        & \beststd{86.92}{.05} & \best{3} \\
    SynFlow
        & \std{81.05}{.05}
        & \std{81.01}{.37}
        & \std{87.92}{.07}
        & \std{88.35}{.04}
        & \std{91.21}{.14}
        & \std{85.47}{.75}
        & \std{85.83}{.16} & 0 \\
    SNIP
        & \std{81.21}{.01}
        & \std{81.62}{.74}
        & \std{88.31}{.12}
        & \std{88.58}{.04}
        & \std{91.32}{.53}
        & \std{86.11}{.40}
        & \std{86.19}{.06} & 0 \\
    FORCE
        & \std{81.31}{.09}
        & \std{79.91}{.74}
        & \std{88.31}{.08}
        & \std{88.46}{.04}
        & \std{91.44}{.23}
        & \std{85.62}{.48}
        & \std{85.84}{.02} & 0 \\
    Taylor-FO
        & \std{81.21}{.01}
        & \std{81.62}{.74}
        & \std{88.31}{.12}
        & \std{88.58}{.04}
        & \std{91.32}{.53}
        & \std{86.11}{.40}
        & \std{86.19}{.06} & 0 \\
    \midrule
    GRaSP
        & \std{81.36}{.14}
        & \std{81.25}{.61}
        & \std{88.11}{.03}
        & \std{88.52}{.12}
        & \std{91.40}{.28}
        & \std{85.69}{.35}
        & \std{86.05}{.20} & 0 \\
    Fisher-Info
        & \std{74.43}{.15}
        & \std{80.39}{.61}
        & \std{80.63}{.64}
        & \std{86.81}{.03}
        & \std{87.50}{.91}
        & \beststd{87.59}{.38}
        & \std{72.31}{.45} & 1 \\
    \bottomrule
    \end{tabular}}
    \caption{%
        Comparing the salience metrics
        on OPT-125m
        with 0.35\% trainable parameters
        on various GLUE tasks,
        following the same format as \Cref{tab:glue_opt350m_bert}.
    }\label{tab:glue_opt125m}
\end{table*}
\begin{table}[!t]
    \centering
    \adjustbox{max width=\linewidth}{%
    \begin{tabular}{l|ccccc|cc}
    \toprule
    \textbf{Method}
        & \textbf{MRPC}   & \textbf{QNLI}
        & \textbf{SST-2}  & \textbf{STS-B}
        & \textbf{QQP}    & \textbf{Avg.}   & \# \\
    \midrule
    \midrule
    LoRA
        & 83.33 & \textbf{92.48}
        & 95.99 & 89.03
        & 89.97 & 90.16 & 1 \\
    \midrule
    Magnitude
        & 77.45 & 90.43
        & 95.18 & 80.33
        & \textbf{90.41} & 86.76 & 1 \\
    Gradient
        & \textbf{87.25} & 92.11
        & 95.53 & \textbf{90.30}
        & 89.02 & \textbf{90.84} & \textbf{2} \\
    SynFlow
        & 78.68 & 90.85
        & \textbf{96.10} & 81.66
        & 88.56 & 87.17 & 1 \\
    SNIP
        & 83.82 & \textbf{92.48}
        & 75.23 & 89.44
        & 85.93 & 85.38 & 1 \\
    FORCE
        & 83.58 & 92.39
        & 89.56 & 88.83
        & 88.31 & 88.53 & 0 \\
    Taylor-FO
        & 83.82 & \textbf{92.48}
        & 75.23 & 89.44
        & 85.93 & 85.38 & 1 \\
    \midrule
    GRaSP
        & 84.80 & 92.46
        & 87.96 & 89.54
        & 88.09 & 88.57 & 0 \\
    Fisher-Info
        & 81.37 & 90.74
        & 83.26 & 84.86
        & 85.27 & 85.10 & 0 \\
    \bottomrule
    \end{tabular}}
    \caption{%
        Comparing the salience metrics
        on OPT-1.3b
        with 0.18\% trainable parameters
        on a subset of the GLUE benchmark,
        following the same format as \Cref{tab:glue_opt350m_bert}.
    }\label{tab:glue_opt1.3b}
\end{table}
\begin{table}[!t]
    \centering
    \cvspace{-.5em}
    \adjustbox{max width=0.9\linewidth}{%
    \begin{tabular}{l|cc|cc|c}
        \toprule
        \textbf{Method}
            & \textbf{HumanEval}
            & \textbf{MBPP}
            & \textbf{Avg.} \\
        \midrule
        \midrule
        LoRA
            & 40.85 & 48.8
            & 44.83 \\
        PiSSA
            & 38.41 & 48.0
            & 43.21 \\
        \midrule
        Gradient
            & 48.78 & 50.0
            & 49.39 \\
        SynFlow
            & 40.85 & 49.0
            & 44.93 \\
        Magnitude
            & 39.02 & 49.2
            & 44.11 \\
        SNIP
            & 46.95 & 51.0
            & 48.98 \\
        FORCE
            & 46.95 & 50.4
            & 34.91 \\
        Taylor-FO
            & 49.39 & 49.4
            & 49.40 \\
        \midrule
        GRaSP
            & 46.34 & 49.8
            & 48.07 \\
        Fisher-Info
            & 46.34 & 47.2
            & 46.77 \\
        \bottomrule
    \end{tabular}}
    \cvspace{-1em}
    \caption{%
        Comparing the salience metrics
        on Llama3-8b with 0.67\%
        trainable parameters.
        We fine-tuned the model
        on CodeFeedback
        and evaluated on HumanEval and MBPP.
        For reference,
        we provide LoRA and PiSSA as baselines
        with the same number of trainable parameters.
    }\label{tab:code_generation}
    \end{table}

\section{Additional Experimental Results}\label{app:results}

\Cref{tab:glue_roberta,tab:glue_opt125m,tab:glue_opt1.3b}
provide additional respective results on GLUE tasks
for the OPT-125m and OPT-1.3b variants,
and BERT-base-uncased.
\Cref{tab:code_generation} shows the results
on HumanEval and MBPP benchmarks
for Llama3-8b model.

\section{Additional Ablation Studies}\label{app:ablation}

We fine-tuned the Gemma2-2b model
on the MetaMathQA dataset
and evaluated it on the GSM8K\_cot task (5-shot)
using flexible extract and strict match metrics.
In order to explore the efficiency-performance trade-off,
we varied for LoRA $ r $ from 4 to 128
and compare it against SPEFT methods
with the same trainable parameters for each config.
The LoRA $ \alpha $ was always kept the same as $ r $.

With the same numbers of training parameters,
LoRA and SPEFT would use almost identical FLOPs per step,
as the added overheads of both
are of the same magnitude
and much smaller (<0.5\% in all of our main experiments)
than the base model.
There was no noticeable difference between LoRA and SPEFT
in terms of computational and memory footprint for all runs.

\begin{figure}[h]
	\centering%
    \begin{subfigure}[b]{\linewidth}
        \centering
        \includegraphics[
            trim=5 20 5 20, clip, width=\linewidth, interpolate=false
        ]{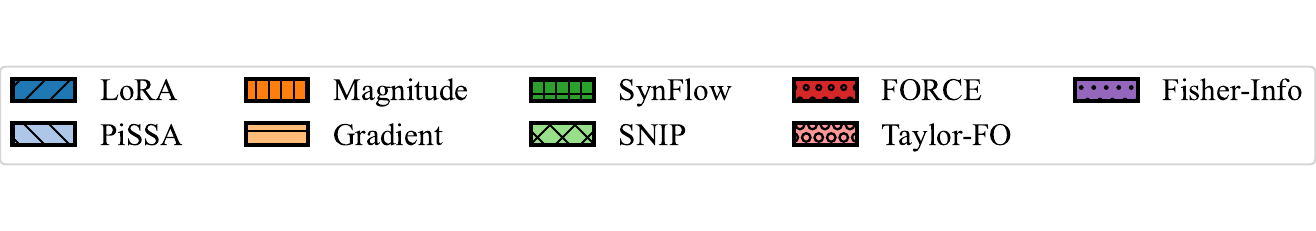}
    \end{subfigure} \\
    \begin{subfigure}[b]{\linewidth}
        \includegraphics[
            width=\linewidth, interpolate=false
        ]{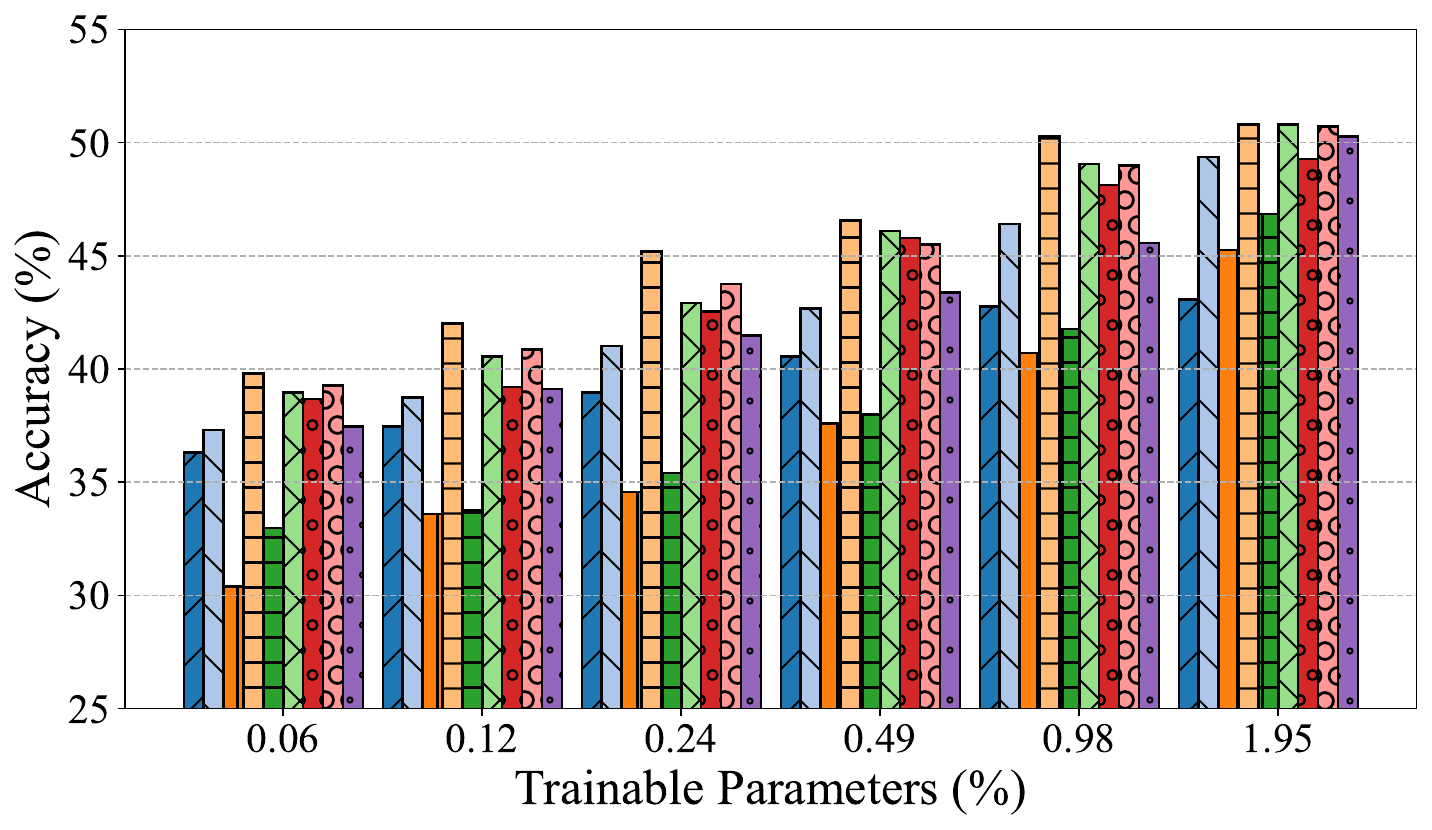}
        \caption{Flexible Extract.}
    \end{subfigure} \\
    \begin{subfigure}[b]{\linewidth}
        \includegraphics[
            width=\linewidth, interpolate=false
        ]{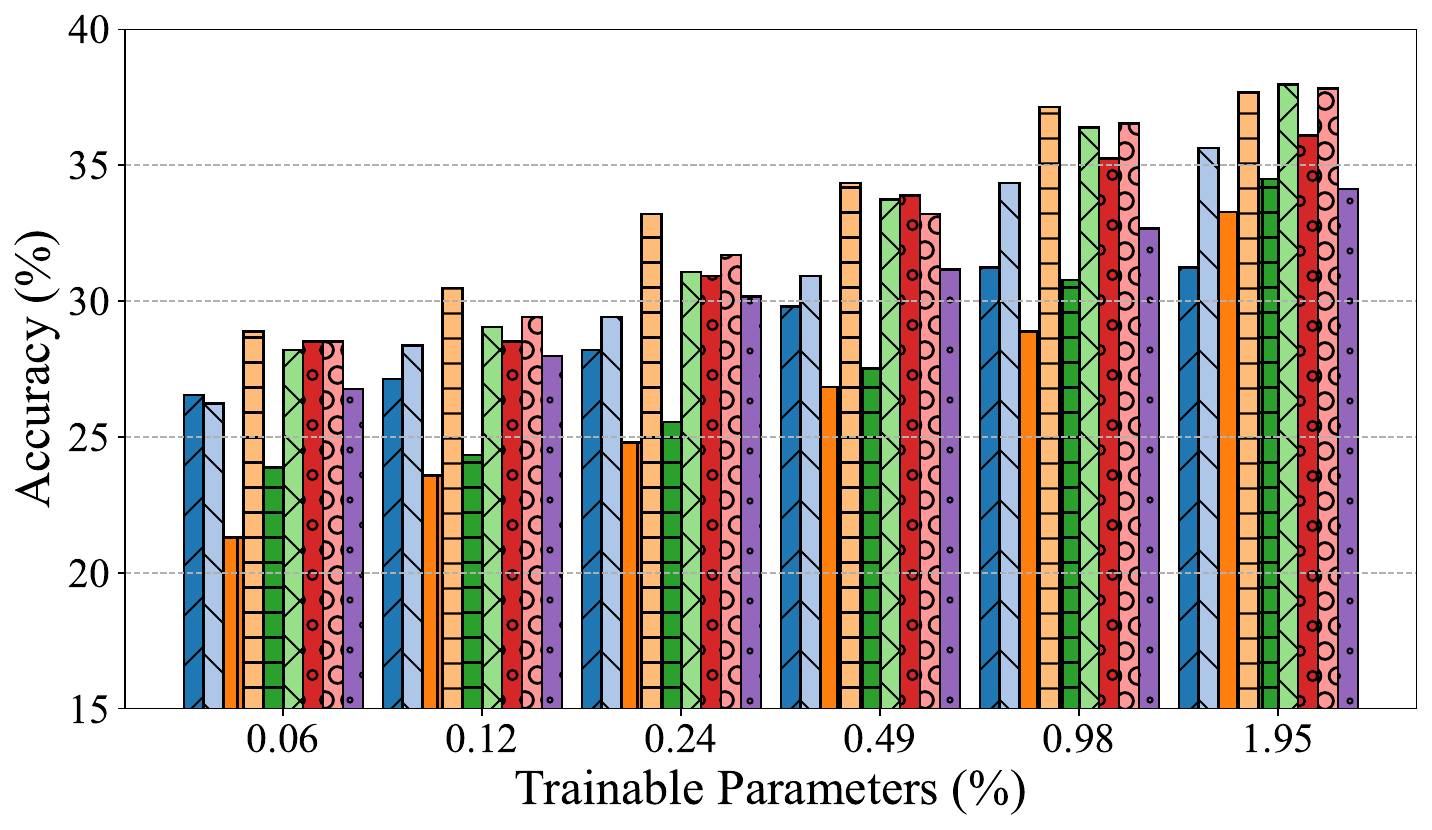}
        \caption{Strict Match.}
    \end{subfigure}
    \caption{%
        Varing the number of trainable parameters
        on Gemma2-2b and GSM8K\_cot (5-shot)
        with LoRA, PiSSA and SPEFT methods.
        The x-axis represents the percentage of trainable parameters,
        while the y-axis denotes accuracy.
    }\label{fig:app_ablation}
\end{figure}

As is shown in \Cref{fig:app_ablation},
the performance of SPEFT methods improve
with increasing trainable parameters
while LoRA results are mostly constant
with increased parameter budget.
Overall,
the gradient-based SPEFT outperformed LoRA
using fewer trainable parameters,
but also widens the gap further as the budget increases.

\section{Computational Resources}
We performed all experiments
on a cluster of NVIDIA A100 40GB GPUs.
The experiments took around 486 GPU-hours
for a single model
on all GLUE subsets
and all salient metrics.
Besides,
it took around 40 GPU-hours
for a single model on Alpaca or OASST2 training
on all low-rank and sparse PEFT methods.
It also took around 80 GPU-hours
to train with all methods
on MetaMath for GSM8k evaluation.
We also spent around 500 GPU-hours
aligning the baseline results with the literature
and determining fine-tuning hyperparameters.

\end{document}